\documentclass[lettersize,journal]{IEEEtran}
\usepackage{amsmath,amsfonts}
\usepackage{algorithmic}
\usepackage{algorithm}
\usepackage{array}
\usepackage[caption=false,font=normalsize,labelfont=sf,textfont=sf]{subfig}
\usepackage{textcomp}
\usepackage{stfloats}
\usepackage{url}
\usepackage{verbatim}
\usepackage{graphicx}
\usepackage{cite}
\usepackage{multirow}
\usepackage{xcolor}

\begin{document}

\title{Two-Stage Framework for Efficient UAV-Based Wildfire Video Analysis with Adaptive Compression and Fire Source Detection}

\author{Yanbing~Bai,~\IEEEmembership{Member,~IEEE,}
Rui-Yang~Ju,~\IEEEmembership{Student Member,~IEEE,}
Lemeng~Zhao, 
Junjie~Hu, Jianchao~Bi, Erick~Mas, 
Shunichi~Koshimura\textsuperscript{\dag}

\thanks{This paper is an extended version of a paper presented at the 2024 IEEE/RSJ International Conference on Intelligent Robots and Systems (IROS), held on October 14–18, 2024, in Abu Dhabi, United Arab Emirates.}
\thanks{Yanbing~Bai and Lemeng~Zhao are with the Center for Applied Statistics, School of Statistics, Renmin University of China, Beijing 100872, China.}
\thanks{Rui-Yang~Ju is with the Graduate School of Informatics, Kyoto University, Kyoto 606-8501, Japan.}
\thanks{Junjie~Hu is with the Shenzhen Institute of Artificial Intelligence and Robotics for Society, Shenzhen 518129, China.}
\thanks{Jianchao~Bi is with the Gaoling School of Artificial Intelligence, Renmin University of China, Beijing 100872, China.}
\thanks{Erick~Mas and Shunichi~Koshimura are with the International Research Institute of Disaster Science, Tohoku University, Sendai 980-8572, Japan.}
\thanks{\textsuperscript{\dag}Corresponding author: Shunichi~Koshimura (koshimura@tohoku.ac.jp).}
%\thanks{Manuscript received xxx xx, 2025; revised xxx xx, 2025.}
}

\markboth{IEEE Journal of Selected Topics in Applied Earth Observations and Remote Sensing}%
{Bai \MakeLowercase{\textit{et al.}}: Two-Stage Framework for Efficient UAV-Based Wildfire Video Analysis with Adaptive Compression and Fire Source Detection}


\maketitle

\begin{abstract}
Unmanned Aerial Vehicles (UAVs) have become increasingly important in disaster emergency response by facilitating aerial video analysis.
Due to the limited computational resources available on UAVs, large models cannot be run efficiently for on-board analysis.
To overcome this challenge, we propose a lightweight and efficient two-stage framework for wildfire monitoring and fire source detection on UAV platforms.
Specifically, in Stage~1, we utilize a policy network to identify and discard redundant video clips, thereby reducing computational costs.
We also introduce a station point mechanism that incorporates future frame information within the sequential policy network to improve prediction accuracy. 
This mechanism allows Stage~1 to operate in a near-real-time manner.
In Stage~2, for frames classified as containing fire, we apply an improved YOLOv8 model to accurately localize the fire source in real-time on selected frames.
We evaluate Stage~1 using the FLAME and HMDB51 datasets, and Stage~2 using the Fire \& Smoke Detection Dataset.
Experimental results show that our method significantly reduces computational costs while maintaining classification accuracy in Stage~1, and achieves high detection accuracy with real-time inference in Stage~2.
\end{abstract}


\section{Introduction}
Forest fires pose a serious threat to both human life and the ecological environment, and have attracted growing attention in recent years~\cite{bouguettaya2022review}.
In particular, UAV-based platforms have become increasingly popular in disaster monitoring and response due to their ability to capture ground information efficiently~\cite{ozkan2021optimization, nagasawa2021model,rashkovetsky2021wildfire}.
UAV-based platforms not only accelerate data acquisition but also reduce the risks involved in manual inspection, thereby enhancing safety in forest fire management.

In recent years, several studies have utilized visual cues such as flames~\cite{goyal2020yolo} and smoke~\cite{alexandrov2019analysis, hossain2020forest, zhan2021pdam} for wildfire detection.
While these methods are feasible, performing fire source detection across an entire video is computationally inefficient, especially given the low probability of wildfire occurrence in any given frame.
A more efficient strategy is to first detect the presence of fire in UAV-based video streams, and only localize the fire source once fire is confirmed.

Therefore, building on our adaptive clip-aware compression and frame selection method proposed in the conference paper~\cite{zhao2024streamlining}, we extend this work into a practical application by introducing a lightweight and efficient two-stage framework.
In Stage~1, we perform frame selection on UAV-based video streams to identify frames containing potential wildfire activity.
In Stage~2, for frames classified as containing fire, we apply an improved YOLOv8 (You Only Look Once, Version 8) model to accurately localize the fire source.

The proposed method in Stage~1 focuses on video understanding, a topic of increasing interest in the field of video streaming data analysis~\cite{bouguettaya2022review}.
Real-time processing of video streams in disaster response scenarios presents significant computational challenges~\cite{koshimura2023digital}.
To address this, researchers have explored methods to reduce computational costs, such as the development of lightweight architectures~\cite{piergiovanni2022tiny, tran2019video, liu2021light} and the selection of representative frames~\cite{guo2022deepcore, meng2020ar, bai2024towards}.
Inspired by these advancements, we propose a novel method in Stage~1: 
the adaptive clip-aware compression and frame selection network.
Unlike traditional methods~\cite{goyal2020yolo, lin2019smoke}, which analyze videos on a frame-by-frame basis, our method selectively combines multiple clips into a single frame.
The decision of when to merge clips is guided by video features, enabling more informed choices. 
By allocating computational resources to information-rich frames and discarding less relevant clips, our method effectively reduces computational costs while maintaining classification accuracy. 

Aerial videos~\cite{koshimura2012tsunami,koshimura2019remote} are often captured under challenging conditions, including blurred frames and abrupt changes caused by unpredictable camera vibrations from air currents. 
To address these issues, we propose a station point mechanism that incorporates features from future frames, allowing the policy network to make more informed and stable decisions during aerial video processing.
This mechanism is a key component of our frame selection module, where the policy network determines the importance of each frame. 
This module assigns scores to individual frames and selects the highest-scoring ones to reconstruct a concise video. 
The process requires no complex pre-processing or post-processing, resembling the editing of a long video to create a shorter, more informative summary. 
The resulting video preserves critical temporal and spatial information, essential for effective disaster monitoring. 

For frames classified as containing fire, we process them further in Stage~2. 
YOLOv8~\cite{jocher2023yolo} is a widely used lightweight object detection model that is well-suited for UAV platforms.
We improve the YOLOv8 architecture by integrating attention mechanisms to enhance the accuracy of fire source localization.

The main contributions of this work are as follows:
\begin{itemize}
\item[(a)] We propose a lightweight and efficient two-stage framework for wildfire monitoring and fire source detection. 
This framework efficiently processes aerial videos by first performing frame selection and classification (Stage~1), and then localizing the fire source on positive frames (Stage~2).
\item[(b)] Stage~1 focuses on UAV-based wildfire video understanding and frame selection, operating in a near-real-time manner. 
We introduce a station point mechanism that leverages future frame information, enhancing the model’s efficiency, robustness, and classification accuracy compared to the baseline method.
\item[(c)] Stage~2 performs fire source detection in real-time on selected frames, integrating attention mechanisms into the YOLOv8 model to achieve higher detection accuracy.
\end{itemize}

The rest of this paper is organized as follows: 
Section~\ref{related} reviews related research on video understanding, frame selection, object detection, and attention modules. 
Section~\ref{method} introduces our proposed two-stage framework and the details of each stage. 
Section~\ref{experiment} describes our experimental setup and presents the experimental results and analysis for both Stage~1 and Stage~2 of our proposed method. 
Section~\ref{discussion} discusses the integration of the attention mechanisms within the Temporal Shift Module (TSM) framework.
Finally, Section~\ref{conclusion} summarizes the main findings of this work and outlines directions for future work.

\section{Related Work}
\label{related}
\subsection{Video Understanding and Frame Selection}
Efficient video understanding models based on representative frames provide feasible solutions to the high computational costs associated with video analysis~\cite{lin2022ocsampler}.
Techniques for frame selection can be broadly categorized into serial (sequential) sampling and parallel sampling.

In serial sampling, frames are processed in temporal order, with a learned policy determining which frame to process next.
AdaFrame~\cite{wu2019adaframe} utilized a memory-augmented Long Short-Term Memory (LSTM) network to retain both temporal and spatial information, using hidden states to guide frame selection.
VideoIQ~\cite{sun2021dynamic} reduced computational overhead by dynamically allocating variable bits or adjusting input resolution on a per-frame basis.

In contrast, parallel sampling processes frames or video clips independently and aggregates the results.
SCSampler~\cite{korbar2019scsampler} scored fixed-length clips and synthesized predictions accordingly.
OCSampler~\cite{lin2022ocsampler} applied a simple one-step reinforcement learning strategy to directly select a comprehensive set of features for video-level modeling. 

Other approaches focus on different criteria for selection.
SMART~\cite{gowda2021smart} utilized a hierarchical selection method, first computing spatial classification scores with a single-frame selector, followed by a global selector that captured both temporal and spatial category information from frame pairs.
MGSampler~\cite{zhi2021mgsampler} focused on motion salience by utilizing two types of motion representations (motion-sensitive and motion-uniform) to effectively distinguish salient motion frames from the background.

In addition, recent methods have introduced novel methods for temporal representation.
For instance, AdaFuse~\cite{meng2021adafuse} dynamically fused channels from current and past feature maps to improve temporal modeling.
However, these methods primarily focused on general video understanding tasks and were typically evaluated on standard benchmark datasets such as ActivityNet~\cite{caba2015activitynet}, FCVID~\cite{jiang2017exploiting}, and Kinetics~\cite{kay2017kinetics}.
As a result, these methods have not been evaluated for UAV-based wildfire videos, which present unique challenges such as sparse informative frames, dynamic smoke patterns, and complex backgrounds. 

\subsection{Object Detection}
For object detection tasks, You Only Look Once (YOLO) is one of the most popular lightweight models due to its balance between model size and accuracy. 
With the introduction of YOLOv8~\cite{jocher2023yolo}, it has been widely adopted in various object detection applications~\cite{ju2023fracture,ju2024global}, including fire detection~\cite{yang2023efficient,talaat2023improved}. 
Specifically, Casas \emph{et al.}~\cite{casas2023assessing} evaluated the performance of YOLOv5, YOLOv6, YOLOv7, YOLOv8, and YOLO-NAS in detecting smoke and wildfires using the Foggia dataset. 
Similarly, Ramos \emph{et al.}~\cite{ramos2025study} evaluated YOLOv8, YOLOv9, and YOLOv10 on the Fire and Smoke dataset for wildfire and smoke detection. 
These works did not propose improvements to the YOLO model architecture, despite evaluating several models from the YOLO series. 

\subsection{Attention Mechanism}
The attention mechanism is an effective method to improve the performance of YOLO series models. 
Hu \emph{et al.}~\cite{hu2018squeeze} introduced Squeeze-and-Excitation Networks (SE-Nets), which significantly improved model performance by generating channel weights through global average pooling, fully connected layers, and Sigmoid functions. 
Woo \emph{et al.}~\cite{woo2018cbam} combined channel and spatial attention to propose the Convolutional Block Attention Module (CBAM), further enhancing the representation capability of Convolutional Neural Networks (CNNs).  
Wang \emph{et al.}~\cite{wang2020eca} proposed Efficient Channel Attention (ECA), which improved performance with fewer parameters by focusing on channel relationships.  
In contrast, Zhang \emph{et al.}~\cite{zhang2021sa} introduced Shuffle Attention (SA), which grouped channel features and integrated complementary features using a Shuffle Unit, achieving excellent results with low model complexity.  
These attention modules can be applied to various neural network architectures to enhance performance.

\begin{figure*}[t] 
\centering 
\includegraphics[width=\linewidth]{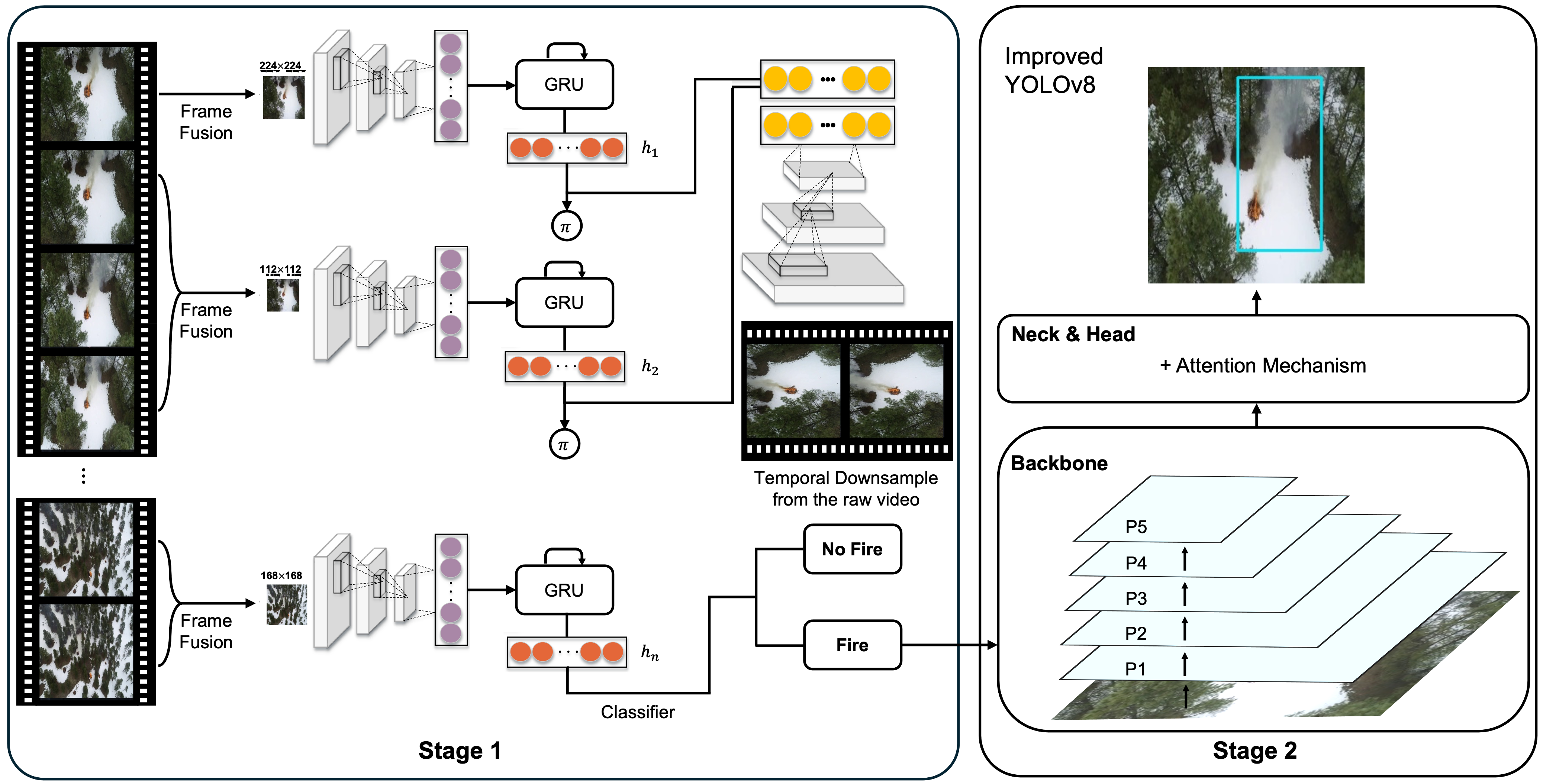}
\caption{Pipeline of the proposed two-stage framework. 
In Stage~1, frame selection is performed through a dynamic policy-based process guided by $\pi$.
Our method enables efficient video understanding by compressing uninformative clips and utilizing extracted features to select informative frames for downstream analysis. 
In Stage~2, the improved YOLOv8 models are employed to accurately localize the fire source.}
\label{fig:overall}
\end{figure*}

\section{Method}
\label{method}
\subsection{Two-Stage Framework}
The pipeline of the proposed two-stage framework is shown in Fig.~\ref{fig:overall}.
In Stage~1, given a video $v$ composed of a sequence of frames $X = {{x_t}}_{t=1}^{N}$, where $N$ denotes the total number of frames and $t$ represents the current time step, the processing begins at $t = 1$.
At the initial step, our method samples multiple station points from the video stream to serve as anchors. 
We extract their features using a 2D-CNN, denoted as ${{s_m}}$.
These station point features provide predictive future context for the policy network $\pi$.

At each subsequent time step $t = i$, the policy network $\pi$ receives as input the concatenation of clip-level features from the feature extraction network $f_s$ and the corresponding station point features.
Based on this input, $\pi$ evaluates the importance of upcoming frames and determines the number of frames to compress at $t = i + 1$.

After the policy network $\pi$ processes the video, it assigns an importance score to each frame. 
Based on these scores, the classifier $f_c$ performs video classification.
This dynamic frame selection process, guided by the pre-trained policy $\pi$, effectively estimates the importance of frames in the stream.

In Stage~2, object detection is performed on these frames classified as containing wildfire.
We employ an improved YOLOv8 model, integrated with attention mechanisms, to accurately localize the fire source within these frames. 

\subsection{Stage~1}
\subsubsection{Frame Fusion}
\begin{figure*}[t] 
\centering 
\includegraphics[width=\linewidth]{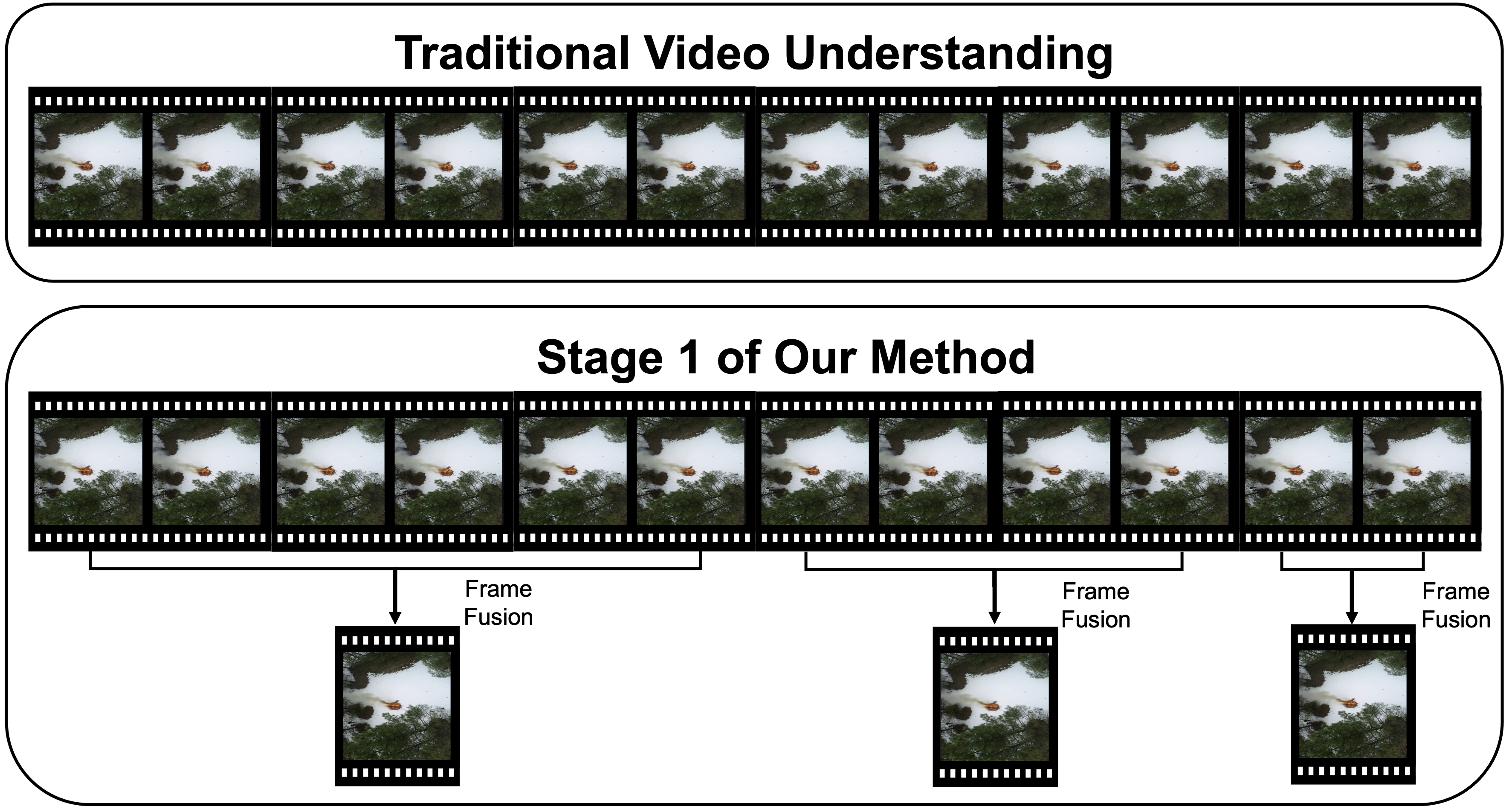}
\caption{Comparison between the traditional method and our method for video understanding.
The traditional method analyzes videos frame by frame, while our method employs Frame Fusion to compress video clips by leveraging video features.}
\label{fig:fusion}
\end{figure*}

To improve computational efficiency while preserving essential video content, we introduce a method named Frame Fusion, as shown in Fig.~\ref{fig:fusion}.
Video classification typically does not require dense frame-level annotations. 
Therefore, for input clips containing multiple frames, our frame fusion method fuses the first and last frames to generate a single representative frame, denoted as $x_i^{\text{fused}}$.
This design is motivated by the observation that frames within a short clip tend to be visually similar, allowing the merged frame to serve as an effective proxy for the entire clip, thereby reducing redundant processing.

\subsubsection{Feature Extraction Network}
At each time step $t = i$, the input frame $x_i^{\text{fused}}$ is first processed by a CNN (e.g., MobileNetV2~\cite{sandler2018mobilenetv2}) to extract spatial features, denoted as $c_i$. 
These features are then fed into an RNN (e.g., GRU) to model temporal dependencies.
The hidden state of the RNN at time step $i$, $h_i$, captures the accumulated video information up to that point.

Formally, this recurrent process is defined as:
\begin{align}
c_i &= \text{CNN}(x_i^{\text{fused}}), \\
h_i &= \text{RNN}(c_i, h_{i-1}),
\end{align}
where $h_{i-1}$ is the hidden state from the previous time step. 
This combined architecture is represented as $f_s$, and its output $h_i$ is subsequently utilized by the policy network $\pi$ for decision making.

\subsubsection{Policy Network}
The policy network $\pi$ determines the number of frames to be fused at the next time step by selecting an action $k_i$ from a discrete action space $A$.
Upon receiving the hidden state $h_i$ from the feature extractor $f_s$ at time $t = i$, $\pi$ identifies the nearest future station point feature $s_m^{\text{nearest}}$ from the set ${{s_m}}$.

The hidden state and the selected station point feature are concatenated and fed into $\pi$, which outputs a predicted probability distribution $a_i^p$ over the action space $A$:
\begin{equation}
a_i^p = \text{softmax}\big(\pi([h_i : s_m^{\text{nearest}}])\big). \label{eq:policy}
\end{equation}

The action space is defined as $A = \{1, 3, 5, 7\}$.
Selecting an action $k_i \in A$ at time $t = i$ means that $k_i$ frames will be merged into a single frame via Frame Fusion at time $t = i + 1$. 
A smaller value of $k_i$ indicates that the current clip is more informative and thus requires higher temporal resolution.

To further optimize resource allocation, frames of varying importance levels are resized to different resolutions. 
We define the resolution set as $R = \{224, 168, 112, 84\}$, where each resolution corresponds to a specific action in $A$. 
This mapping allows $\pi$ to dynamically adjust the level of feature detail based on predicted importance. 
For instance, if the policy determines that current features are sufficient, it may merge five or seven frames at a lower resolution; conversely, when finer details are needed, it merges several frames at a higher resolution.

The architecture of $\pi$ consists of a Group Normalization~\cite{wu2018group} layer followed by a fully connected layer. 
The use of Group Normalization accelerates convergence during training by stabilizing feature distributions.

\subsubsection{Frame Selection}
The policy network $\pi$ effectively evaluates the importance of video frames, as reflected by its fusion decisions:
the greater the number of frames fused (a larger $k_i$), the less important they are considered. 
To quantify this, we introduce a preference score $S$ to evaluate the significance of each frame.

According to Eq.~(\ref{eq:policy}), at time step $t = i$, the policy network outputs a probability distribution over the action space $A$, $a_i^p = \{a_{i1}^p, ..., a_{i\|A\|}^p\}$.
For the clip to be fused at $t = i + 1$, denoted as $clip_{i+1}$, we define its preference score $S_{i+1}$ as the inverse expected action:
\begin{equation}
S_{i+1} = \sum_{j=1}^{\|A\|} \frac{a_{ij}^p}{A_j}, \label{eq:selection}
\end{equation}
where $A_j$ is the $j$-th action in $A$ (e.g., $A_1=1, A_2=3, ...$). 
This score $S$ can also be calculated using the Gumbel-Softmax ($a_i^{gs}$) or Gumbel-Max ($a_i^{gm}$) outputs in place of $a_i^p$.

To further refine the preference score $S$ at the frame level, the central frame of the $clip_{i+1}$ is assigned the full score $S_{i+1}$, with a gradual decay of 10\% applied symmetrically toward the neighboring frames.

\subsubsection{Gumbel Softmax Trick}
The action space $A$ is discrete, which poses challenges for optimization via gradient backpropagation. 
In such cases, reinforcement learning is commonly employed to train the policy network. 
However, reinforcement learning often suffers from slow convergence across a wide range of applications~\cite{wu2019liteeval}. 

To address this limitation, we adopt the Gumbel Softmax trick~\cite{jang2017categorical}, which enables direct optimization of the policy network through gradient-based methods.
The Gumbel Softmax is an effective reparameterization technique that transforms previously non-differentiable categorical distributions into differentiable approximations, introducing stochasticity during training and encouraging exploration. 

Specifically, the policy network selects actions using the Gumbel Max operation rather than the non-differentiable $\arg\max$~\cite{jang2017categorical}:
\begin{equation}
a^{gm}_i = \arg\max \left( \log(a^p_i) + G_i \right), \label{eq:gumbel_max}
\end{equation}
where $G_i$ is a vector of independent and identically distributed Gumbel noise, calculated as $G_{ij} = -\log(-\log(U_{ij}))$ with $U_{ij} \sim \text{Uniform}(0,1)$.

For backpropagation, a differentiable Gumbel-Softmax approximation is used to approximate the $\arg\max$ operation and enable gradient computation:
\begin{equation}
a_i^{gs} = \text{softmax}\left( \frac{\log(a^p_i) + G_i}{\tau} \right), \label{eq:gumbel_softmax}
\end{equation}
where $\tau$ is the softmax temperature hyperparameter. 
When $\tau > 0$, the output is a smooth distribution; as $\tau \to 0$, it converges to one-hot vectors. 
Following prior works~\cite{meng2020ar, sun2021dynamic}, we initialize $\tau$ at 5 and gradually anneal it to 0 over the course of training.

\subsection{Stage~2}
\subsubsection{Baseline Model}
The architecture of the baseline model (i.e., YOLOv8) consists of a Backbone, Neck, and Head, as illustrated in Fig.~\ref{fig:overall}.
The Backbone adopts the Cross Stage Partial (CSP) structure~\cite{wang2020cspnet}, which reduces computational cost while improving the learning capability of the CNNs.

Compared to YOLOv5~\cite{jocher2020yolov5}, YOLOv8 introduces a new Coarse-to-Fine (C2f) module, replacing the previously used Concentrated-Comprehensive Convolution (C3) module.
C2f integrates the C3 design with the Extended ELAN (E-ELAN) concept~\cite{wang2022designing} from YOLOv7~\cite{wang2023yolov7}, and comprises two convolutional blocks connected by multiple bottlenecks, each implemented as a Convolution–BatchNorm–SiLU (CBS).

For the Neck architecture, YOLOv8 retains the Feature Pyramid Network (FPN)~\cite{lin2017feature} and Path Aggregation Network (PAN)~\cite{liu2018path} structure adopted from YOLOv5. 
The FPN enhances low-level features via up-sampling, while the PAN strengthens high-level localization through down-sampling. 
This combination improves the model's ability to detect objects across multiple scales. 
In addition, YOLOv8 removes redundant convolutional operations during the up-sampling stage to further improve inference efficiency.

In the Head design, YOLOv8 adopts a decoupled architecture, separating classification and regression branches to improve task-specific learning performance.
Moreover, the Objectness branch is removed, leaving only classification and regression outputs.
YOLOv8 also transitions from an anchor-based method~\cite{ren2015faster} to an anchor-free model~\cite{duan2019centernet}, which directly predicts the object centroid and its boundary offsets for localization.

\subsubsection{Attention Module}
Ju \emph{et al.}~\cite{ju2024yolov8,chien2025yolov8} proposed YOLOv8-AM and applied it to pediatric wrist fracture detection, achieving impressive results.
In this work, we adopt these improved YOLOv8 models to wildfire images captured by UAVs and further extend them to enable localization of the fire sources. 
Specifically, we employ three variants: YOLOv8+ResCBAM (ResBlock + CBAM), YOLOv8+ECA, and YOLOv8+SA.

As illustrated in Fig.~\ref{fig:overall}, we integrate the attention modules into the Neck of YOLOv8.
In particular, one attention module is added after each of the four C2f modules, with the specific module selected from ResCBAM, ECA, or SA, depending on the variant.

Convolutional Block Attention Module (CBAM)~\cite{woo2018cbam} is a lightweight and effective attention mechanism composed of two sequential sub-modules: channel attention and spatial attention.
It enhances feature representation by adaptively refining both the channel-wise and spatial-wise focus of the network.
CBAM is well-suited for wildfire detection, where precise localization of small and irregular fire regions is essential.
By sequentially applying channel and spatial attention, CBAM enables YOLOv8 to better capture fire-related features, such as subtle color or texture variations, while suppressing irrelevant background noise often present in the aerial images.
Integrating CBAM into the Neck of YOLOv8 allows the model to adaptively enhance the most discriminative intermediate features before detection, improving the accuracy of fire source localization.

Efficient Channel Attention (ECA) module~\cite{wang2020eca} is a lightweight attention mechanism designed to capture essential inter-channel dependencies while minimizing computational and parameter overhead.
Unlike traditional attention mechanisms that rely on dimensionality reduction, which may lead to information loss, ECA employs a 1D convolutional operation to model local cross-channel interactions after global average pooling.
For wildfire source localization based on UAV images, accurate detection of fire regions requires heightened sensitivity to subtle variations in color, brightness, and texture, features often encoded in the channel dimension.
By integrating ECA into the Neck of YOLOv8, the model can more effectively highlight channel-wise features that are relevant to fire patterns, while suppressing less informative responses.

Shuffle Attention (SA) module~\cite{zhang2021sa} is an efficient attention mechanism that integrates both channel and spatial attention through feature grouping and channel shuffling.
This module fuses the enhanced channel and spatial features and redistributes them through channel shuffling, enabling effective cross-group information exchange and holistic feature refinement.
Wildfire scenes often exhibit complex spatial patterns, such as scattered flames, diffuse smoke, and varying vegetation textures, which necessitate the joint modeling of channel-wise semantics and spatial locality.
By applying SA in the Neck of YOLOv8, the model gains the ability to capture fine-grained regional cues and their broader contextual relationships.

\subsection{Training Procedure}
The training process begins by sampling video data and extracting features using the combined CNN-RNN backbone.
The policy network $\pi$ then dynamically determines the optimal number of frames to compress via Frame Fusion.

To improve training stability and efficiency, we adopt a multi-step training procedure:
\textbf{Step 1: Pre-train CNN.} 
We first fine-tune the CNN feature extractor (e.g., MobileNetV2) using frame-level labels, while all other modules remain frozen.
\textbf{Step 2: Pre-train RNN.} 
We freeze the pre-trained CNN and train the GRU using clip-level labels. 
In this step, the policy network is bypassed, and frames are fed to the GRU using a default uniform sampling strategy.
\textbf{Step 3: Jointly Train Policy and Classifier.} 
We freeze both the pre-trained CNN and GRU modules. Then, we jointly train the policy network ($\pi$) and the final classifier ($f_c$).

During \textbf{Step 3}, only the parameters of $\pi$ and $f_c$ are updated by minimizing the composite loss function, which combines classification, balanced, and FLOPs-aware terms.
This strategy approximates end-to-end optimization of the full model while ensuring stable convergence and efficient resource usage.
The resulting framework enables effective and adaptive video understanding, making it well-suited for UAV-based wildfire monitoring applications.

\begin{figure*}[t] 
\centering 
\includegraphics[width=\linewidth]{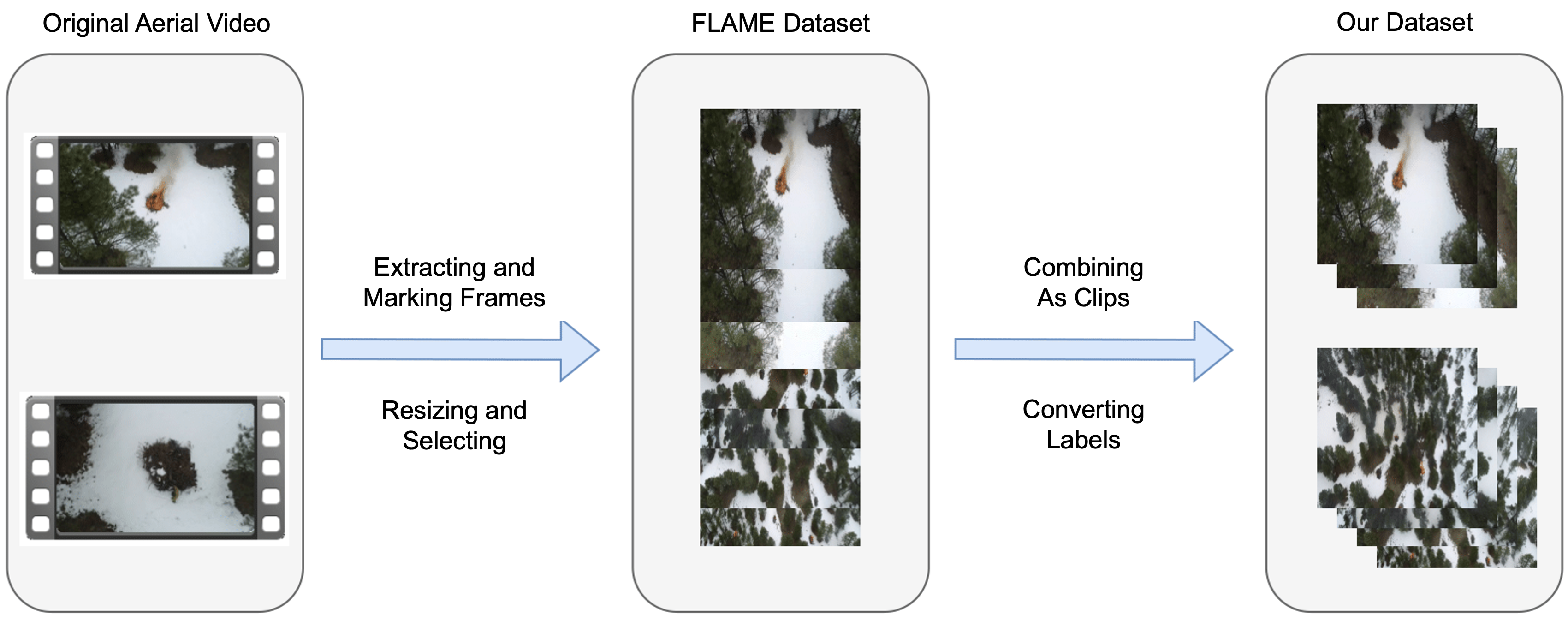}
\caption{The pipeline illustrating the construction process of FLAME wildfire video clips for classification tasks.}
\label{fig:data_process}
\end{figure*}

\begin{table*}[t]
\centering
\caption{Detailed usage statistics for the three datasets used in this work.}
\setlength{\tabcolsep}{12pt}
\begin{tabular}{llcccc}
\hline
\textbf{Dataset} & \textbf{Task} & \textbf{Classes} & \textbf{Training Set} & \textbf{Validation Set} & \textbf{Test Set} \\
\hline
FLAME~\cite{shamsoshoara2021aerial} & Video Classification & 2 (Fire / Non-Fire) & 39,375 Frames & - & 8,617 Frames \\
HMDB51~\cite{kuehne2011hmdb} & Video Classification & 51 & 3,570 Clips & 1,530 Clips & 1,666 Clips \\
Fire \& Smoke Detection~\cite{akhtamov2023fire} & Object Detection & 2 (Fire / Smoke) & 11,035 Images & 3,260 Images & 3,266 Images \\
\hline
\end{tabular}
\label{tab:dataset}
\end{table*}

\subsection{Loss Function}
\subsubsection{Stage~1}
We design three loss components to train the model in Stage~1. 
The total loss $L$ is defined as a weighted sum:
\begin{equation}
L = L_c + \beta L_b + \gamma L_g, \label{eq:loss}
\end{equation}
where $\beta$ and $\gamma$ are hyperparameters that balance the contributions of each loss term.
In our experiments, we set $\beta = 0.3$ and $\gamma = 0.1$, following the settings used in VideoIQ~\cite{sun2021dynamic}.

The first component, cross-entropy loss $L_c$, measures the classification error between the predicted output and the ground truth:
\begin{equation}
L_c = \mathbb{E}\left[-Y \log \bigl(f_c(h_{\text{final}})\bigr)\right], \label{eq:Lc}
\end{equation}
where $h_{\text{final}}$ denotes the final hidden state from the temporal aggregation module (e.g., GRU), and $f_c$ is the final classification layer.

To prevent the policy network from converging to a trivial solution, we introduce a balance loss $L_b$. 
This loss penalizes the deviation of the action distribution from a uniform distribution using the squared L2 norm:
\begin{equation}
L_b = \sum_{k \in A} \left( p_k - \frac{1}{\|A\|} \right)^{2}, \label{eq:Lb}
\end{equation}
where $p_k = \mathbb{E} \left[ \frac{1}{T} \sum_{t=1}^T \mathbb{I}(k_t = k) \right]$ represents the average frequency (empirical probability) of selecting action $k$. 
Here, $k_t \in \{1, 3, 5, 7\}$ is the action selected at time step $t$, $T$ denotes the total number of time steps, and $\mathbb{I}(\cdot)$ is the indicator function.

To promote computational efficiency while preserving accuracy, we also include a Giga Floating-point Operations (GFLOPs) loss $L_g$, which penalizes the average computational cost:
\begin{equation}
L_g = \mathbb{E} \left[ \frac{1}{T} \sum_{t=1}^T \text{GFLOPs}(k_t) \right], \label{eq:Lg}
\end{equation}
where $\text{GFLOPs}(k_t)$ is the computational cost associated with executing action $k_t$. 
This $L_g$ term encourages the policy to favor actions with lower computational cost, balancing accuracy and efficiency.

\subsubsection{Stage~2}
We adopt the loss functions used in YOLOv8~\cite{jocher2023yolo}. 
Specifically, the Binary Cross-Entropy (BCE) loss is applied to the classification branch, while the Distribution Focal Loss (DFL)~\cite{li2020generalized} and Complete Intersection over Union (CIoU) loss~\cite{zheng2021enhancing} are employed for the regression branch.

The BCE loss function is defined as:
\begin{equation}
\text{Loss}_{\text{BCE}} = -w\left[y_n \log x_n + (1 - y_n) \log(1 - x_n)\right], \label{eq:bce}
\end{equation}
where $w$ is a weighting factor, $y_n$ denotes the ground truth label, and $x_n$ represents the predicted value of the model.

The Distribution Focal Loss (DFL) is defined as follows: 
Given a continuous target value $y$ that falls between two discrete bins $y_n$ and $y_{n+1}$, with corresponding predicted probabilities $S_n$ and $S_{n+1}$ (where $S_n + S_{n+1} = 1$), the loss is defined as follows:
\begin{equation}
\begin{split}
\text{Loss}_{\text{DFL}} = & - \left[ (y_{n+1} - y) \log(S_n) + (y - y_n) \log(S_{n+1}) \right]. \label{eq:dfl}
\end{split}
\end{equation}

In addition, the Complete Intersection over Union (CIoU) loss is based on the Distance-IoU (DIoU) loss~\cite{zheng2020distance} by introducing an additional penalty term that considers the aspect ratio consistency between the predicted bounding box and the ground truth. 
The CIoU loss is shown as follows:
\begin{equation}
\text{Loss}_{\text{CIoU}} = 1 - \text{IoU} + \frac{d^2}{c^2} + \frac{v^2}{(1 - \text{IoU}) + v}, \label{eq:ciou}
\end{equation}
where $d$ is the Euclidean distance between the centers of the predicted and ground-truth boxes, and $c$ is the diagonal length of the smallest enclosing box covering both boxes. 
The term $v$ measures the consistency of the aspect ratio and is defined as:
\begin{equation}
v = \frac{4}{\pi^2} \left( \arctan \frac{w_{\text{gt}}}{h_{\text{gt}}} - \arctan \frac{w_p}{h_p} \right)^2, \label{eq:v}
\end{equation}
where $w_p$ and $h_p$ represent the width and height of the predicted bounding box, and $w_{\text{gt}}$ and $h_{\text{gt}}$ represent those of the ground-truth bounding box.

\begin{figure*}[t] 
\centering 
\includegraphics[width=\linewidth]{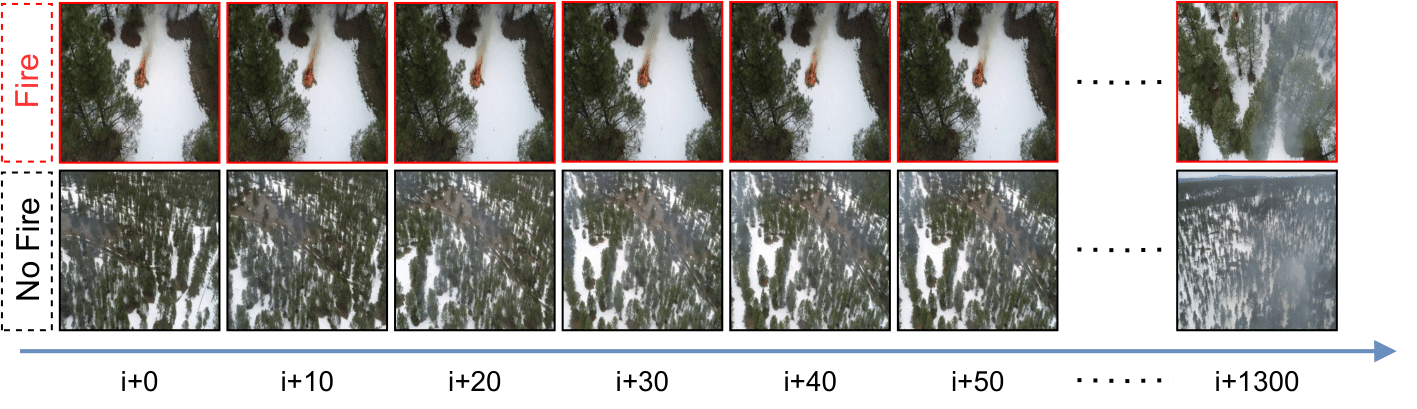}
\caption{A representative example illustrating the two types of labels used in the FLAME dataset~\cite{shamsoshoara2021aerial}. 
The arrow indicates the temporal order of frames within a video sequence.}
\label{fig:dataset}
\end{figure*}

\section{Experiments}
\label{experiment}
\subsection{Dataset}
\subsubsection{FLAME Dataset}
The FLAME (Fire Luminosity Airborne-based Machine learning Evaluation) dataset~\cite{shamsoshoara2021aerial} is a publicly available collection of wildfire images and videos captured by UAVs. 
For the video classification task, we utilize the seventh and eighth subsets of this dataset.
As illustrated in Fig.~\ref{fig:data_process}, each video sample is defined as a sequence of 64 consecutive frames extracted from the same video.
If any frame within the sequence contains wildfire, the entire sample is labeled as positive.
This procedure yields a curated dataset of 615 training clips and 134 testing clips. 
These clips total 483 positive (wildfire) and 266 negative (wildfire-free), resulting in an approximate 2:1 ratio. 
A representative sample from this video dataset is shown in Fig.~\ref{fig:dataset}, and the detailed usage statistics are shown in TABLE~\ref{tab:dataset}.

\subsubsection{HMDB51 Dataset}
The HMDB51 (A large video database for human motion recognition) dataset~\cite{kuehne2011hmdb} is a widely used open-domain video recognition benchmark consisting of 6,766 video clips collected from diverse real-world sources.
It includes 51 distinct action categories, each containing a minimum of 101 clips.
In this study, we use HMDB51 (split 1) as an auxiliary dataset to evaluate the effectiveness of our method in efficient video recognition.
The detailed usage statistics for this dataset are shown in TABLE~\ref{tab:dataset}.

\subsubsection{Fire \& Smoke Detection Dataset}
Since the FLAME and HMDB51 datasets do not include manually annotated wildfire localization, we utilize the Fire \& Smoke Detection Dataset~\cite{akhtamov2023fire} to evaluate the performance of our method in fire source localization.
This dataset, published by Azimjon Akhtamov, an AI researcher at Chungbuk National University (CBNU), South Korea, contains over 17,000 images covering fire and smoke in various environments, including urban, forest, industrial, and indoor settings, and the detailed usage statistics for this dataset are shown in TABLE~\ref{tab:dataset}.
It is well-suited for tasks related to early wildfire detection and Earth hazard monitoring.

\subsection{Baseline}
\subsubsection{Stage~1}
In Stage~1 of our proposed framework, we compare the performance of our frame selection method against three baselines: random frame selection, uniform frame selection, and the state-of-the-art (SOTA) method SMART~\cite{gowda2021smart}. 
In random selection, frames are sampled randomly across the entire video, while uniform selection samples frames at even intervals. 
For all three baseline methods, once the frames are selected, the final prediction is obtained by average pooling the outputs of a common backbone applied to each selected frame.

\subsubsection{Stage~2}
In Stage~2 of our method, we use the original YOLOv8~\cite{jocher2023yolo} and YOLOv9~\cite{wang2024yolov9} as the baseline models and compare their performance with our improved variants: YOLOv8+ECA, YOLOv8+SA, and YOLOv8+ResCBAM.

\begin{table}[t]
\centering
\caption{Quantitative comparison (Accuracy) of Temporal Shift Module (TSM) models trained with different numbers of selected frames, evaluated on the FLAME dataset~\cite{shamsoshoara2021aerial}.}
\setlength{\tabcolsep}{18pt}{
\begin{tabular}{c|c|c|c}
\hline
\multirow{2}{*}{Method} & \multicolumn{3}{c}{Number of Frames} \\ \cline{2-4}
 & 8 & 20 & 28 \\ \hline
Uniform & \textbf{87.3\%} & 88.1\% & 84.3\% \\
Random & 85.1\% & 88.1\% & 86.6\% \\
SMART & 84.3\% & 88.1\% & 88.8\% \\
Ours & \textbf{87.3\%} & \textbf{91.0\%} & \textbf{91.0\%} \\ \hline \noalign{\smallskip}
\multicolumn{4}{l}{The highest accuracy in each group is highlighted in \textbf{bold}.} \\
\end{tabular}}
\label{tab:tsm_acc}
\end{table}

\subsection{Implementation Details}
\subsubsection{Stage~1}
We set the number of station points to 2.
To improve computational efficiency, we adopt MobileNetV2~\cite{sandler2018mobilenetv2} as the backbone for the feature extractor $f_s$, which is also used to process the station points.
The Gated Recurrent Unit (GRU)~\cite{cho2014learning} has a hidden dimension of 512 and consists of a single layer.

The training pipeline for the Selector consists of three steps:
\textbf{Step 1 (Pre-train CNN)}: 
We first fine-tune the MobileNetV2, pretrained on ImageNet~\cite{deng2009imagenet}, using frame-level labels for 100 epochs. The initial learning rate is set to 0.01 and decays by a factor of 10 at epochs 50, 70, and 90.
\textbf{Step 2 (Pre-train RNN)}: 
We freeze the pre-trained MobileNetV2 and train the GRU using clip-level labels for 20 epochs with an initial learning rate of 1.45e-5. In this step, a uniform sampling strategy is used.
\textbf{Step 3 (Jointly Train Policy)}: 
Finally, both MobileNetV2 and the GRU are frozen, and the policy network ($\pi$) and classifier ($f_c$) are trained jointly using the composite loss (Eq.~\ref{eq:loss}) with an initial learning rate of 0.01.

\subsubsection{Stage~2}
For Stage~2 of our method, all models were trained using the PyTorch framework.
Experiments were conducted on an NVIDIA GeForce RTX 4090 GPU, with the batch size set to 16 to accommodate memory constraints.

For training hyperparameters, we follow the configuration used by Ju \emph{et al.} and Chien \emph{et al.}~\cite{ju2024yolov8,chien2025yolov8,ju2026pediatric,chien2024yolov9}, employing Stochastic Gradient Descent (SGD)~\cite{ruder2016overview} as the optimizer.
The weight decay is set to 5e-4, with a momentum of 0.937.
The initial learning rate is set to 1e-2, and training is performed for 100 epochs.

\begin{figure}[t]
\centering
\includegraphics[width=\linewidth]{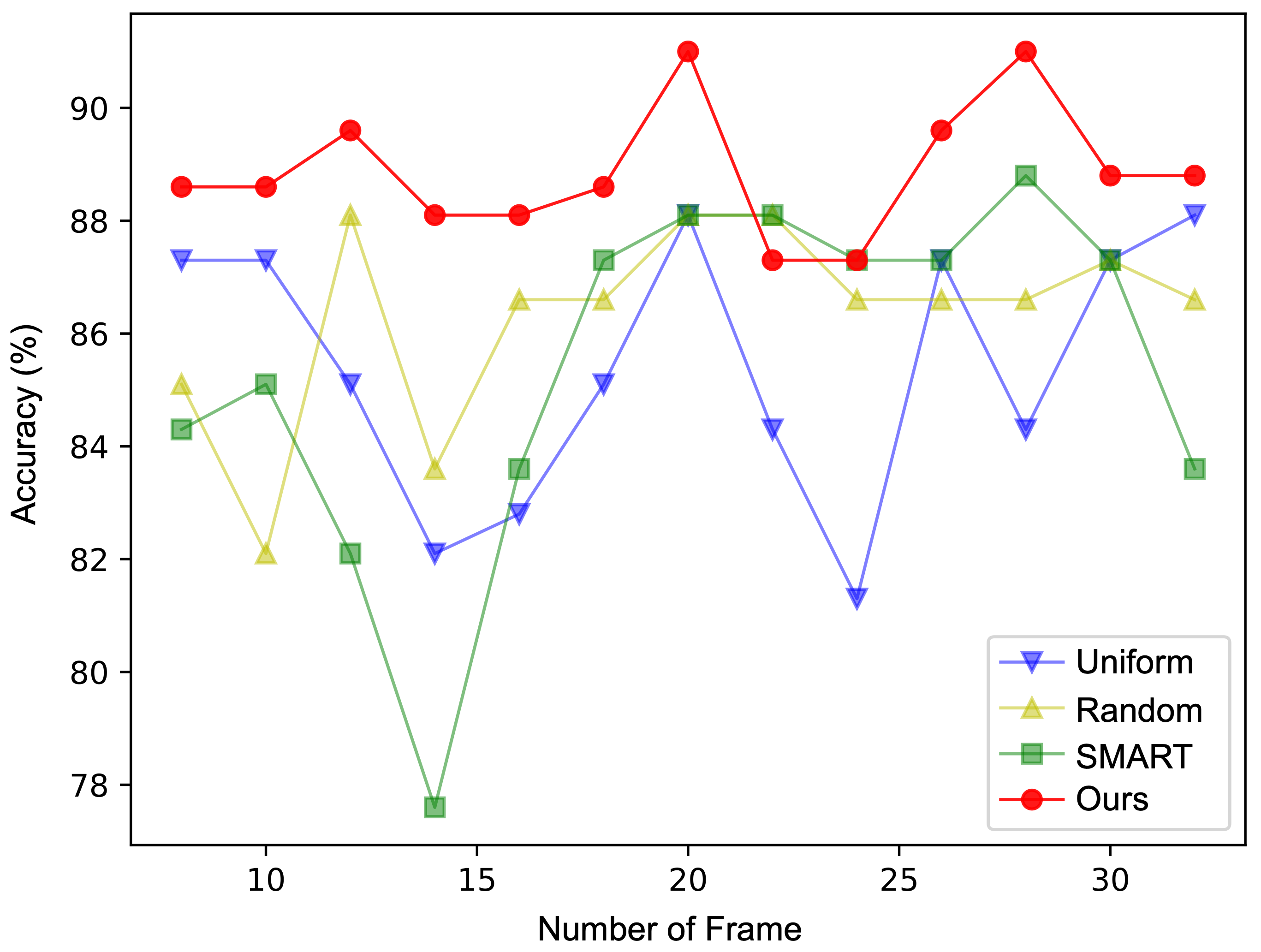}
\caption{Comparison of baseline methods (Uniform and Random), the SOTA method SMART~\cite{gowda2021smart}, and our method across different numbers of selected frames, evaluated on the FLAME dataset~\cite{shamsoshoara2021aerial}.}
\label{fig:quantitative}
\end{figure}

\subsection{Evaluation Metrics}
\subsubsection{Stage~1}
For Stage~1 of our method, classification accuracy is consistently used as the primary evaluation metric for the video classification task. Accuracy is defined as the ratio of correctly predicted samples to the total number of samples and is computed as:
\begin{equation}
\text{Accuracy} = \frac{TP + TN}{TP + TN + FP + FN},
\end{equation}
where $TP$, $TN$, $FP$, and $FN$ represent true positives, true negatives, false positives, and false negatives, respectively.

In addition to accuracy, we evaluate the computational cost of each model in terms of floating-point operations (FLOPs).

\subsubsection{Stage~2}
The F-score is a widely used metric for evaluating the performance of object detection models, as it incorporates both precision and recall to provide a balanced assessment of accuracy.
The F-score is computed using the following equation:
\begin{equation}
\text{F-Score} = \frac{\left(1+\beta^{2}\right) \times \text{Precision} \times \text{Recall}}{\beta^{2} \times \text{Precision} + \text{Recall}}, \label{f-score}
\end{equation}

When $\beta$ = 1, the F-score becomes the F1-score, which assigns equal weight to precision and recall and is interpreted as their harmonic mean.
The F1-score is calculated as follows:
\begin{equation}
\begin{split}
\text{F1-Score} =&\frac{2 \times Precision \times Recall}{Precision + Recall} \\
&=\frac{2T_P}{2T_P+F_P+F_N}, \label{f1-score}
\end{split}
\end{equation}
where $T_P$, $F_P$, and $F_N$ denote true positives, false positives, and false negatives, respectively.

Mean Average Precision (mAP) is a widely used evaluation metric in object detection tasks, providing a comprehensive measure of a model’s performance across all classes. 
In this work, we adopt mAP@50 as the evaluation metric, where mAP@50 refers to the mean of the average precision for each class computed at an Intersection over Union (IoU) threshold of 0.5.

In addition, we report the inference time of each model, measured in milliseconds (ms). 
Inference time refers to the total duration required by the YOLO models to process a wildfire frame, from input through pre-processing, model inference, and post-processing, until the final prediction is obtained.

\subsection{Quantitative Comparison for Stage~1}
The selected frames are used to train the Temporal Shift Module (TSM)~\cite{lin2019tsm}, a state-of-the-art (SOTA) classification model, to evaluate the effectiveness of our frame selection method.
The corresponding results are summarized in TABLE~\ref{tab:tsm_acc}.

As shown in TABLE~\ref{tab:tsm_acc}, our method achieves superior performance by maintaining high classification accuracy using only 8 frames, outperforming even the results obtained from processing the full video.
In addition, our method significantly outperforms the baseline methods (Uniform and Random) as well as the SOTA method SMART~\cite{gowda2021smart} when using the same number of frames. 

These results indicate that our method effectively selects the most representative frames, enabling more efficient and accurate classification of wildfire videos.

\subsection{Analysis of the Behavior of Frame Selection}
As illustrated in Fig.~\ref{fig:quantitative}, additional experiments are conducted across a wider range of frame counts to assess whether the observed results are affected by data randomness due to the limited dataset size.
The results show that our method consistently outperforms the other three frame selection methods.

Across nearly all frame counts, our method achieves the highest classification accuracy, maintaining performance within the 88–91\% range, which highlights both robustness and stability.
In contrast, the performance of the other methods exhibits greater variability, particularly SMART and Random, which suffer substantial accuracy drops when the number of selected frames is small (e.g., 14 frames).

Among all methods, our method exhibits the smoothest performance curve, reflecting strong generalization capabilities with respect to varying frame numbers.
By comparison, the other methods show pronounced fluctuations, with the Random method showing the highest sensitivity to the number of selected frames.

\begin{table}[t]
\centering
\caption{Ablation study of video classification performance (Accuracy in \%) on the FLAME~\cite{shamsoshoara2021aerial} and HMDB51~\cite{kuehne2011hmdb} datasets.}
\setlength{\tabcolsep}{1.4pt}{
\begin{tabular}{c|cccc|cccc}
\hline
\multirow{3}{*}{Method} & \multicolumn{4}{c|}{FLAME} & \multicolumn{4}{c}{HMDB51} \\ \cline{2-9}
 & \begin{tabular}[c]{@{}c@{}}Test\\ Acc.\end{tabular} & \begin{tabular}[c]{@{}c@{}}Train\\ Acc.\end{tabular} & \begin{tabular}[c]{@{}c@{}}FLOPs\\ /Video\end{tabular} & \begin{tabular}[c]{@{}c@{}}FLOPs\\ /Frame\end{tabular} & \begin{tabular}[c]{@{}c@{}}Test\\ Acc.\end{tabular} & \begin{tabular}[c]{@{}c@{}}Train\\ Acc.\end{tabular} & \begin{tabular}[c]{@{}c@{}}FLOPs\\ /Video\end{tabular} & \begin{tabular}[c]{@{}c@{}}FLOPs\\ /Frame\end{tabular} \\ \hline
GRU & 79.10\% & \textbf{98.61\%} & 26.28G & 0.411G & \textbf{40.72\%} & \textbf{95.78\%} & 30.19G & 0.32G \\
GRU+$\pi$ & \textbf{82.84\%} & 97.40\% & \textbf{2.79G} & \textbf{0.044G} & 38.50\% & 92.34\% & \textbf{1.29G} & \textbf{0.013G} \\ \hline \noalign{\smallskip}
\multicolumn{9}{l}{The best value in each comparison group is highlighted in \textbf{bold}.} \\
\end{tabular}}
\label{tab:ablation}
\end{table}

\begin{figure}[t]
\centering
\includegraphics[width=\linewidth]{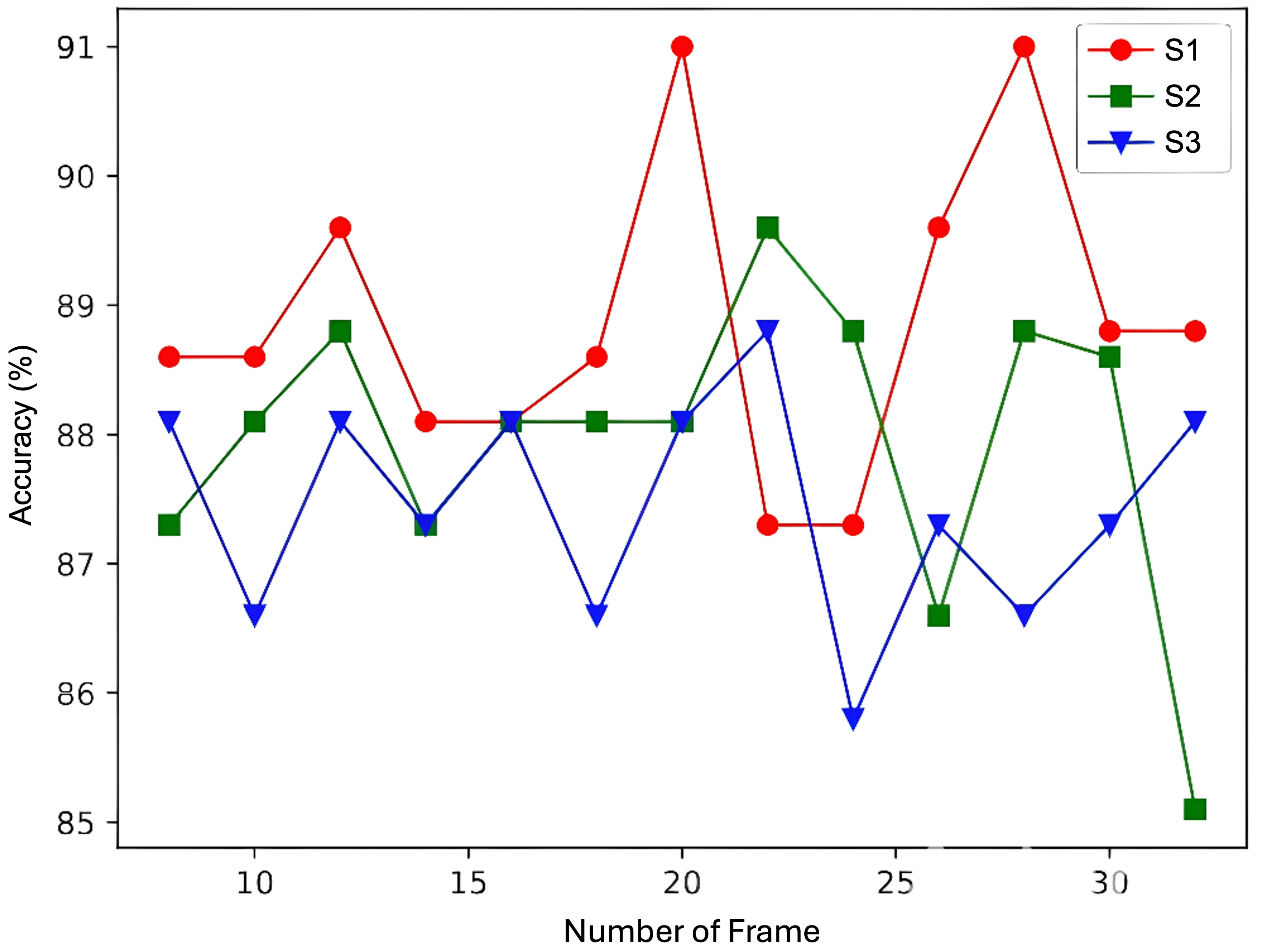}
\caption{Performance comparison of frame selection scoring methods (S1, S2, S3). 
These scores correspond to the preference score $S$ (Eq.~\ref{eq:selection}) calculated using three different policy network outputs: 
$a_i^{gm}$ (Eq.~\ref{eq:gumbel_max}), 
$a_i^p$ (Eq.~\ref{eq:policy}), 
and $a_i^{gs}$ (Eq.~\ref{eq:gumbel_softmax}), respectively.}
\label{fig:score}
\end{figure}

\begin{table}[t]
\centering
\caption{Performance comparison of video classification with different action space $A$.}
{\setlength{\tabcolsep}{4pt}
\begin{tabular}{c|cccc}
\hline
$A$ & Test Acc. & Train Acc. & FLOPs/Video & FLOPs/Frame\\ \hline
$\left\{1, 3, 5\right\}$ & \textbf{82.84\%} & 97.24\% & 3.41G & 0.053G \\
$\left\{1, 3, 5, 7, 9\right\}$ & 77.60\% & \textbf{97.56\%} & 3.26G & 0.051G \\
$\left\{1, 3, 5, 7\right\}$ & \textbf{82.84\%} & 97.40\% & \textbf{2.79G} & \textbf{0.044G} \\ \hline \noalign{\smallskip}
\multicolumn{5}{l}{$\left\{1, 3, 5\right\}$ corresponds to $R = \left\{224, 168, 112\right\}$.} \\
\multicolumn{5}{l}{$\left\{1, 3, 5, 7, 9\right\}$ corresponds to $R = \left\{224, 168, 140, 112, 84\right\}$.} \\ 
\multicolumn{5}{l}{The best value in each comparison group is highlighted in \textbf{bold}.} \\
\end{tabular}}
\label{tab:action_space}
\end{table}

\begin{table}[t!]
\centering
\caption{Performance comparison of video classification with different numbers of the station points.}
\setlength{\tabcolsep}{9pt}{
\begin{tabular}{c|cccc}
\hline
 & Test Acc.  & Train Acc. & FLOPs/Video & FLOPs/Frame \\ \hline
0 & 75.37\% & \textbf{98.2\%} & 6.61G & 0.103G \\
1 & 80.60\% & 98.1\% & \textbf{2.54G} & \textbf{0.040G} \\
2 & \textbf{82.84\%} & 97.4\% & 2.79G & 0.044G \\ 
3 & 78.36\% & 97.7\% & 3.06G & 0.048G \\ \hline \noalign{\smallskip}
\multicolumn{5}{l}{The best value in each comparison group is highlighted in \textbf{bold}.} \\
\end{tabular}}
\label{tab:station_points}
\end{table}

\subsection{Ablation Study for Stage~1}
To evaluate the effectiveness of the proposed method, we integrate both the feature extraction network $f_s$ and the frame selection policy $\pi$ for video classification, and compare the results with those of a baseline GRU model that relies solely on $f_s$.
The results are summarized in TABLE~\ref{tab:ablation}.

On the FLAME dataset, our method (GRU + policy) not only achieves a notable improvement in test accuracy (82.84\% vs. 79.10\%) but also significantly reduces computational cost.
Specifically, the video-level FLOPs are reduced from 26.28G to 2.79G, and frame-level FLOPs from 0.411G to 0.044G, representing a remarkable reduction.
These results highlight the effectiveness of the learned policy in identifying and focusing on informative frames, thus enhancing both accuracy and efficiency.

To further assess the generalizability of our method, we evaluate it on the HMDB51 dataset.
As shown in TABLE~\ref{tab:ablation}, although the test accuracy decreases slightly (from 40.72\% to 38.50\%), the computational cost is drastically reduced.
Video-level FLOPs drop from 30.19G to 1.29G, and frame-level FLOPs from 0.32G to 0.013G, yielding a 23-fold improvement in efficiency.
These results demonstrate the strong generalizability and scalability of our method across diverse video datasets.

\begin{table*}[t]
\centering
\caption{Quantitative comparison of fire source detection on the Fire \& Smoke Detection Dataset using YOLOv8 and our models with model size \textbf{small} and input image size set to \textbf{1024}.}
\setlength{\tabcolsep}{16pt}{
\begin{tabular}{l|ccccccc}
\hline
Model & Params & FLOPs & F1 & Precision & Recall & mAP@50 & Inference$^1$ \\ \hline
YOLOv8 & 11.14M & 28.6G & 0.63 & 61.8\% & 66.3\% & 65.5\% & 1.8ms \\
YOLOv8+SA & 11.14M & 28.7G & 0.64 & 62.8\% & 65.4\% & 65.2\% & 1.9ms \\
YOLOv8+ECA & 11.14M & 28.7G & 0.64 & 60.4\% & 68.0\% & 66.0\% & 1.8ms \\
YOLOv8+ResCBAM & 16.06M & 38.3G & 0.64 & 60.4\% & 67.9\% & 66.0\% & 2.1ms \\
YOLOv9 & 7.29M & 27.4G & 0.64 & 61.5\% & 67.5\% & 65.5\% & 2.2ms \\ \hline \noalign{\smallskip}
\multicolumn{8}{l}{$^1$Inference means the inference time per image measured using NVIDIA GeForce RTX 4090 GPU.}
\end{tabular}}
\label{tab:yolov8_small}
\end{table*}

\begin{table*}[t]
\centering
\caption{Quantitative comparison of fire source detection on the Fire \& Smoke Detection Dataset using YOLOv8 and our models with model size \textbf{medium} and input image size set to \textbf{1024}.}
\setlength{\tabcolsep}{16pt}{
\begin{tabular}{l|ccccccc}
\hline
Model & Params & FLOPs & F1 & Precision & Recall & mAP@50 & Inference$^1$ \\ \hline
YOLOv8 & 25.86M & 79.1G & 0.64 & 61.0\% & 66.1\% & 65.8\% & 6.1ms \\
YOLOv8+SA & 25.86M & 79.1G & 0.64 & 60.0\% & 68.2\% & 65.9\% & 6.3ms \\
YOLOv8+ECA & 25.86M & 79.1G & 0.64 & 60.0\% & 68.4\% & 66.2\% & 6.1ms \\
YOLOv8+ResCBAM & 33.83M & 98.2G & 0.65 & 63.6\% & 67.2\% & 66.6\% & 6.9ms \\
YOLOv9 & 20.16M & 77.6G & 0.64 & 61.8\% & 66.6\% & 66.0\% & 4.1ms \\ \hline \noalign{\smallskip}
\multicolumn{8}{l}{$^1$Inference means the inference time per image measured using NVIDIA GeForce RTX 4090 GPU.}
\end{tabular}}
\label{tab:yolov8_medium}
\end{table*}

\subsection{Hyperparameter Experiment for Stage~1}
\subsubsection{Effect of Preference Score}
In addition to using the score $a_i^{gm}$ (defined in Eq.~\ref{eq:gumbel_max}) for frame selection, we further investigate two alternative scoring strategies: $a_i^p$ (from Eq.~\ref{eq:policy}) and $a_i^{gs}$ (from Eq.~\ref{eq:gumbel_softmax}).
These scoring methods are referred to as S1, S2, and S3, respectively, corresponding to the preference score $S$ (Eq.~\ref{eq:selection}) calculated using each of these three outputs.
To evaluate their effectiveness, we conduct a series of experiments varying the number of selected frames. 
The results are presented in Fig.~\ref{fig:score}.

As shown in Fig.~\ref{fig:score}, the S1-based method consistently achieves higher accuracy across most configurations compared to S2 and S3.
Notably, S1 demonstrates more stable and superior performance, maintaining accuracy above 88\% in most cases and reaching a peak of over 91\%.
In contrast, S2 and S3 exhibit greater variability and lower peak accuracy.
These results underscore the advantage of using $a_i^{gm}$ for frame scoring.
Therefore, we adopt S1 as the scoring strategy in our proposed model to ensure optimal and reliable performance.

\subsubsection{Effect of Action Space}
We evaluate the impact of different action space configurations $A$ on video classification performance, with the results summarized in TABLE~\ref{tab:action_space}. 
The analysis shows that the configuration $\{1, 3, 5, 7\}$ achieves the best balance, delivering the highest test accuracy (82.84\%) and the lowest computational cost (2.79G FLOPs).
In contrast, expanding the action space to $\{1, 3, 5, 7, 9\}$ resulted in both higher FLOPs and lower test accuracy.

\subsubsection{Effect of Station Point}
We investigate the impact of different numbers of station points on video classification performance, with the results summarized in TABLE~\ref{tab:station_points}.
The policy network exhibits substantial performance differences with and without station points.  
Introducing a single station point yields a 5\% improvement in test accuracy while significantly reducing FLOPs (from 6.61G to 2.54G). 
In contrast, the policy network without station points adopts a more conservative strategy.  
As the number of station points increases, the network tends to fuse more frames.  
Overall, using two station points achieves an optimal balance between computational cost and classification accuracy.

\begin{figure*}[t] 
\centering 
\includegraphics[width=\linewidth]{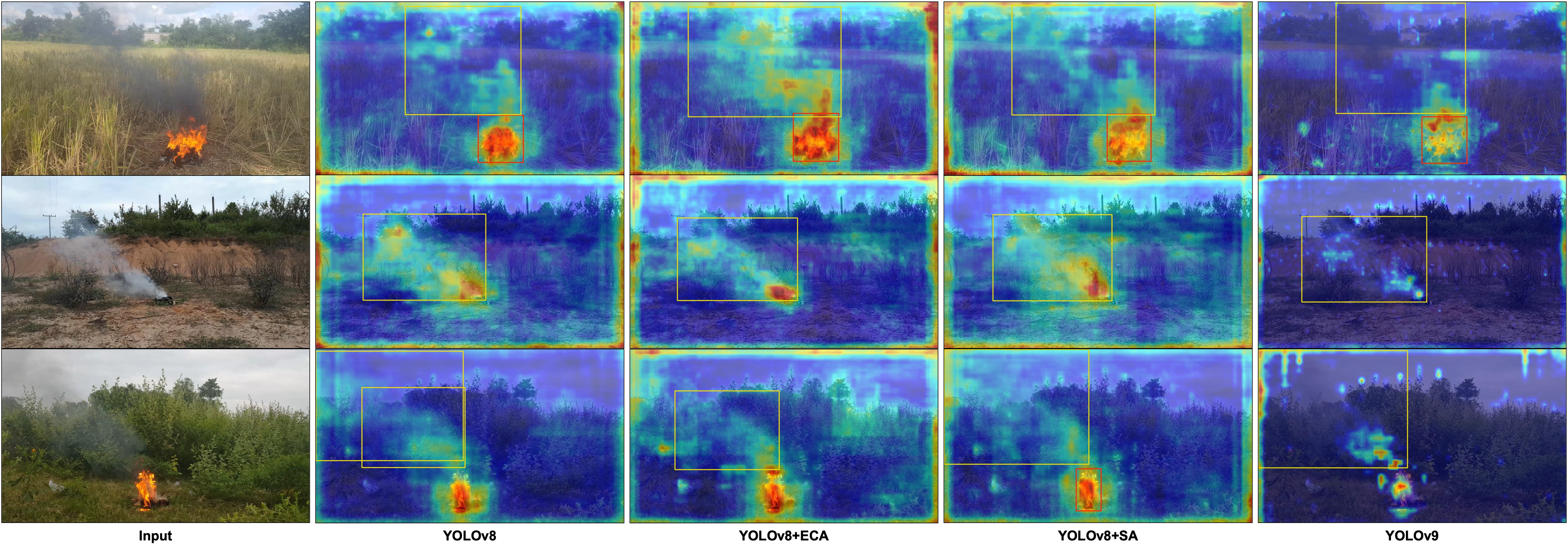}
\caption{Examples of heatmaps obtained by different models for wildfire detection.}
\label{fig:heatmap}
\end{figure*}

\begin{figure*}[t]
\centering
\includegraphics[width=\linewidth]{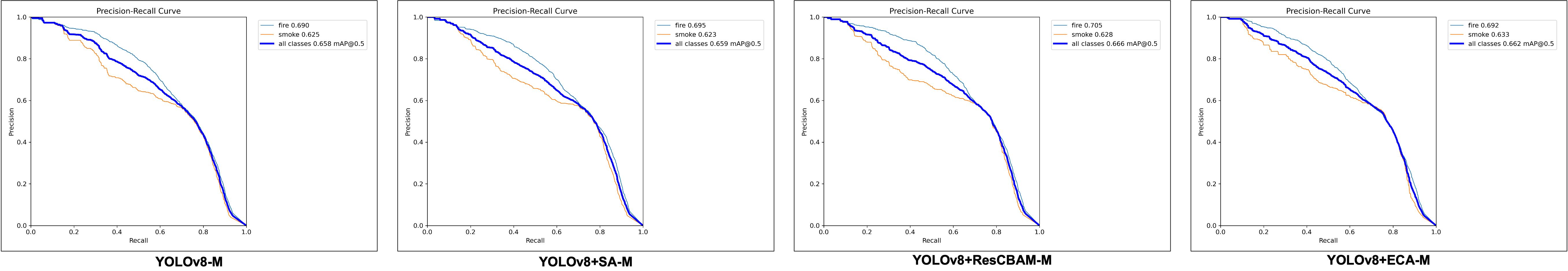}
\caption{Detailed Precision–Recall curves for four different models across each class in the Fire \& Smoke Detection Dataset~\cite{akhtamov2023fire}.}
\label{fig:prcurve}
\end{figure*}

\subsection{Quantitative Comparison for Stage~2}
For the fire source detection task, we set all input image sizes to 1024 to enable a fair comparison between the baseline and the proposed methods. 
We evaluate two model scales: Small (S) and Medium (M), with results presented in TABLEs~\ref{tab:yolov8_small} and~\ref{tab:yolov8_medium}.

In the Small model setting, both YOLOv8+ECA and YOLOv8+ResCBAM outperform the original YOLOv8, which achieves 65.5\% mAP@50.
Specifically, both YOLOv8+ECA and YOLOv8+ResCBAM obtain mAP@50 values of 66.0\%.
Moreover, the inference times remain efficient: 
YOLOv8+ECA matches the baseline similarly with an inference time of 1.8ms, while YOLOv8+ResCBAM suffers only a slight increase to 2.1ms. 

YOLOv8+SA shows minor improvements in F1 and precision but a slight drop in mAP@50 (65.2\%).
In comparison, YOLOv9 achieves comparable accuracy with 65.5\% mAP@50, while reducing the number of parameters to 7.29M. 
However, its inference time of 2.2ms is notably slower than the YOLOv8 baseline (1.8ms).
Considering model complexity and efficiency, YOLOv8+ECA remains the preferred model for fire source detection on resource-limited platforms.

For platforms with greater computational resources, the Medium model version is preferred, as it achieves higher mAP@50 values compared to the Small version.
As shown in TABLEs~\ref{tab:yolov8_small} and~\ref{tab:yolov8_medium}, the mAP@50 of the baseline YOLOv8 increases from 65.5\% (Small) to 65.8\% (Medium), accompanied by a rise in inference time from 1.8ms to 6.1ms. 

All of our model variants demonstrate improved performance.
YOLOv8+ECA achieves 66.2\% mAP@50 while matching the baseline's parameters, FLOPs, and inference time, according to TABLE~\ref{tab:yolov8_medium}.
YOLOv8+ResCBAM achieves the highest accuracy of 66.6\% mAP@50, though with increased complexity: parameters rise from 25.86M to 33.83M, FLOPs from 79.1G to 98.2G, and inference time from 6.1ms to 6.9ms.

YOLOv9 presents a competitive alternative with 66.0\% mAP@50 while maintaining fewer parameters (20.16M) and lower inference time (4.1ms), providing a good trade-off between accuracy and efficiency.
Therefore, if model accuracy is the primary consideration, YOLOv8+ResCBAM-M is the most suitable choice for fire source detection in high-resource computing environments.

\subsection{Visualization for Stage~2}
To further evaluate how different attention modules influence wildfire-related feature extraction, we employ Grad-CAM++~\cite{chattopadhay2018grad} to visualize the attention distribution of each model, as shown in Fig.~\ref{fig:heatmap}. 
All models use the Medium configuration.
In these heatmaps, brighter regions indicate higher importance assigned by the model. The red bounding box corresponds to the ``fire'' class, whereas the yellow bounding box denotes ``smoke''.

In the first row, YOLOv8 correctly activates the fire region but also highlights a large portion of the surrounding sky and vegetation, indicating substantial background sensitivity.
YOLOv8+ECA produces a significantly more concentrated activation pattern, with attention almost exclusively focused on the flames. 
YOLOv8+SA shows similarly focused activation, although with slightly more background response compared to the ECA variant. 
YOLOv9 exhibits a dispersed and noisy activation pattern, with scattered false responses appearing across the sky and vegetation, and weaker localization on the fire region.

In the second row, YOLOv8 generates relatively weak and fragmented activations that only partially cover the smoke source. 
YOLOv8+ECA provides clear and concentrated attention precisely aligned with the smoke region. 
YOLOv8+SA again behaves comparably to YOLOv8+ECA, producing a compact and reliable activation area. 
YOLOv9, in contrast, shows an almost entirely dark heatmap, failing to activate in the smoke region and only producing faint responses over the foreground grass.

In the third row, YOLOv8 detects the fire region but also activates surrounding grass, revealing susceptibility to background elements. 
YOLOv8+ECA yields clean and well-localized activations, with nearly all attention concentrated on the flames. 
YOLOv8+SA also maintains a strong and focused activation pattern, slightly less clean than the ECA variant but still highly localized. 
YOLOv9 produces dispersed activations across the flames, nearby vegetation, and the upper sky region, reflecting poor localization and high background noise.

\subsection{Practical Deployment Considerations}
In real-world deployment scenarios, such as UAV-based wildfire monitoring, both computational resources and environmental conditions constrain model selection. 
Small models, such as YOLOv8+ECA-S, are well-suited for UAVs with limited onboard processing capabilities, enabling real-time detection with minimal latency while maintaining reasonable accuracy. 
Medium models, such as YOLOv8+ResCBAM-M, are more computationally demanding but can be deployed on ground stations or high-performance UAV platforms where maximizing detection accuracy is critical.

In addition, environmental factors, including variable illumination, smoke, and background clutter, can introduce noise and degrade detection performance. 
TABLEs~\ref{tab:yolov8_small} and~\ref{tab:yolov8_medium} indicate that both ECA and ResCBAM modules improve the robustness of YOLOv8 under such conditions, enhancing the reliability of fire source detection. 
Furthermore, the trade-off between inference speed and model complexity should be carefully evaluated in the context of detection urgency, UAV battery constraints, and communication bandwidth.

\begin{table}[t]
\centering
\caption{Quantitative comparison (Accuracy in \%) of CrossFormer-TSM models trained with different numbers of selected frames.}
\setlength{\tabcolsep}{14pt}{
\begin{tabular}{c|c|c|c}
\hline
\multirow{2}{*}{Method} & \multicolumn{3}{c}{Number of Frames} \\ \cline{2-4}
 & 8 & 20 & 28 \\ \hline
Frozen CrossFormer & 84.4\% & 85.3\% & 85.3\% \\
Trained CrossFormer & 86.8\% & 88.2\% & 87.4\% \\ \hline
\end{tabular}}
\label{tab:crossformer}
\end{table}

\subsection{Analysis of the P-R Curve of Fire Source Detection}
The Precision-Recall (P-R) Curve plots Recall on the x-axis and Precision on the y-axis, with each point corresponding to a different confidence threshold, forming a continuous curve. 
We generate P-R Curves for the Medium versions of various models and report fire source detection performance (mAP@50) for two classes: fire and smoke, as well as the overall performance across all classes, as illustrated in Fig.~\ref{fig:prcurve}. 

Among the models, YOLOv8+ResCBAM-M achieves the highest performance in detecting the ``fire'' class, with a mAP@50 of 70.5\%, while YOLOv8+ECA-M performs best in detecting the ``smoke'' class, reaching a mAP@50 of 63.3\%. 
In addition, all P-R Curves exhibit a steep drop, indicating a significant decline in Precision at high Recall levels. 
Compared to fire detection, the performance on smoke detection remains relatively lower, which suggests significant potential for improvement.

\section{Discussion}
\label{discussion}
In Stage~1 of our method, we further investigate the incorporation of attention-based architectures as feature extractors within the TSM framework by replacing conventional CNNs with CrossFormer~\cite{zhang2023crossformer}. 
Two configurations are set as: 
one with the CrossFormer backbone frozen and another with its parameters fully trainable. 
As shown in TABLE~\ref{tab:crossformer}, the trainable CrossFormer consistently outperforms the frozen version, highlighting the benefits of end-to-end optimization across all frame settings.

Nevertheless, the performance gains from this architectural enhancement remain relatively modest. 
Notably, increasing the number of selected frames from 20 to 28 does not yield significant improvements and, in some cases, results in a slight decrease in accuracy.
We attribute this limited gain primarily to the relatively limited size of the FLAME dataset~\cite{shamsoshoara2021aerial}, which constrains the representational capacity of CrossFormer and may lead to underutilization of its attention-based features.

To further enhance model performance, particularly when employing complex attention-based architectures, we posit that training on a larger-scale dataset is essential. 
Such a dataset would provide greater diversity and volume, allowing the model to fully exploit the expressive power of architectures like CrossFormer and potentially achieve more substantial performance improvements within the TSM framework.

\section{Conclusion}
\label{conclusion}
In this work, we propose a lightweight and efficient two-stage framework for wildfire monitoring and detection.
In Stage~1, we introduce a novel method to enhance the efficiency of aerial video understanding in wildfire monitoring. 
In Stage~2, we employ improved YOLOv8 models for accurate fire source localization.

Experiments demonstrate that our method outperforms several baseline methods in Stage~1, providing a high-quality set of selected frames that enables the Temporal Shift Module (TSM) to achieve high accuracy with significantly reduced computational cost.
Notably, the compression mechanism in Stage~1 is model-independent and compatible with various backbone architectures, making it broadly applicable to video-based disaster response scenarios.
In Stage~2, our improved YOLOv8 models can improve mAP@50 performance while maintaining similar levels of model parameters, FLOPs, and inference time.

A major challenge in real-world deployment is the trade-off between false positives and false negatives in wildfire detection. 
False positives can trigger unnecessary alerts, increasing the burden on emergency response systems, while false negatives pose the more serious risk of delayed detection, potentially allowing fires to spread uncontrollably. 
Therefore, a balance between these error types is critical for reliable system operation. 

Potential strategies to address this challenge include fusing multi-sensor data (e.g., UAV-based wildfire videos with satellite imagery), implementing a hierarchical alert mechanism requiring multi-stage confirmation before resource deployment, and introducing human-in-the-loop verification for high-confidence scenarios. 
Such strategies can help the framework maintain high sensitivity to early-stage fires while effectively suppressing false positives in real-world wildfire monitoring tasks.

Notably, Stage~1 of our proposed framework is currently unsuitable for strict real-time deployment because it relies on future frames for stable inference, resulting in delays.
However, this limitation may be less critical in scenarios where a short processing latency is acceptable, such as near-real-time monitoring. 
To meet strict real-time requirements, future work will explore stable inference methods that do not rely on future frames. 
Potential strategies include designing causal or online temporal models that utilize only past and current frames, predicting future frame features from available information, aggregating temporal features within a sliding window to reduce reliance on distant frames, employing incremental updates to minimize computation, and applying teacher-student knowledge distillation to transfer performance from high-accuracy models that use future frames to lightweight real-time models. 

In addition, the current frame fusion strategy, which compresses only two frames per clip, may cause feature loss. 
Future research will focus on developing more advanced frame compression techniques capable of integrating multiple frames while maintaining computational efficiency.

Currently, Stage~1 and Stage~2 in this work are trained and evaluated on independent datasets. 
We acknowledge that the end-to-end cascading performance is not provided because, to the best of our knowledge, there is no publicly available dataset that simultaneously supports wildfire video classification and wildfire detection. 
As part of our future work, we plan to collect and construct a dedicated dataset that enables joint evaluation of both tasks.

\section*{Acknowledgment}
This work was supported by JSPS KAKENHI (Grants-in-Aid for Scientific Research, 21H05001).
This research was supported by Public Computing Cloud, Renmin University of China. 

\bibliographystyle{ieeetr}
\bibliography{ieeeabrv}
\end{document}